\documentclass[letterpaper, 10 pt, conference]{ieeeconf-modified}  

\IEEEoverridecommandlockouts                              

\usepackage[numbers]{natbib}
\usepackage{graphicx}
\graphicspath{ {./figures} }
\usepackage{siunitx}
\usepackage{algorithm}
\usepackage{algorithmic}
\usepackage{amsmath} 
\usepackage{amssymb}  
\usepackage{textcomp}
\usepackage{hyperref}
\usepackage[capitalize]{cleveref}
\usepackage{booktabs}
\usepackage[caption=false]{subfig}
\usepackage{xcolor}

\crefname{equation}{}{}

\newcommand{\gitlink}{https://dlr-alr.github.io/2023-humanoids-completion/}

\newcommand{\ltwo}[1]{\left\lVert#1\right\rVert^2}
\DeclareMathOperator*{\argmin}{arg\,min}

\title{\LARGE \bf
Combining Shape Completion and Grasp Prediction\\ for Fast and Versatile Grasping with a Multi-Fingered Hand
}
\author{Matthias Humt$^{1,2}_*$, Dominik Winkelbauer$^{1,2}_*$, Ulrich Hillenbrand$^1$, and Berthold Bäuml$^{1,3}$
\thanks{$^1$ The author is with the DLR Institute of Robotics and Mechatronics.}
\thanks{$^2$ The author is with the Technical University of Munich (TUM).}
\thanks{$^3$ The author is with the Deggendorf Institute of Technology (DIT).}
\thanks{
    $_*$ Equal contribution. \newline
    Contact: \tt\footnotesize{matthias.humt | berthold.baeuml@dlr.de}
}
}

\begin{document}

\maketitle
\thispagestyle{empty}
\pagestyle{empty}

\begin{abstract}
  Grasping objects with limited or no prior knowledge about them is a highly relevant skill in assistive robotics.
  Still, in this general setting, it has remained an open problem, especially when it comes to only partial observability and versatile grasping with multi-fingered hands.
We present a novel, fast, and high fidelity deep learning pipeline consisting of a shape completion module that is based on a single depth image, and followed by a grasp predictor that is based on the predicted object shape.
  The shape completion network is based on VQDIF and predicts spatial occupancy values at arbitrary query points.
  As grasp predictor, we use our two-stage architecture that first generates hand poses using an autoregressive model and then regresses finger joint configurations per pose.
  Critical factors turn out to be sufficient data realism and augmentation, as well as special attention to difficult cases during training.
      Experiments on a physical robot platform demonstrate successful grasping of a wide range of household objects based on a depth image from a single viewpoint.
  The whole pipeline is fast, taking only about \SI{1}{\second} for completing the object's shape (\SI{0.7}{\second}) and generating 1000 grasps (\SI{0.3}{\second}).\\(\href{\gitlink}{\gitlink})
\end{abstract}

\section{Introduction}

A major step towards real-world autonomy of robot systems in general, and of versatile humanoids in particular, is their ability to grasp objects at a first-time encounter, that is, a successful handling of novel objects. This scenario is likely to occur, e.g., in assistive robotics operating in human living and working environments, or indeed in any environment that is not strongly constrained in terms of objects and tasks (hospitals, elderly care homes, shop floors, offices).

Accordingly, the goal of unknown object grasping has been pursued for more than a decade. In early studies, one main approach was to transfer grasps from a database of known objects to novel ones with similar shapes~\cite{Curtis2008GraspKnowledgeBase,Goldfeder2009ColumbiaGraspDatabase}. However, only the grasp pose and the preshape of the robot hand or gripper were retrieved in this way, without having control over the grasp contacts finally made. Grasp poses were also inferred from probabilistic models~\cite{Song2011GraspGraphicalModelsLatentSpace}, and the preshape from Eigengrasp planners~\cite{Ciocarlie2009HandPostureSubspace}. Going one step further, in~\cite{Hillenbrand2012GraspWarping,Stouraitis2015GraspWarping}, we have warped individual finger contacts onto novel objects, but also being restricted to a high level of shape similarity. All in all, the early approaches were suffering from a lack of generality and control. More recently, there has been progress on grasping of unknown objects from a wider range~\cite{Sundermeyer2021ContactGraspNet}, but being tailored to parallel jaw grippers and hence simple specific grasp types.

\begin{figure}[t!]
\centering
\includegraphics[width=\linewidth]{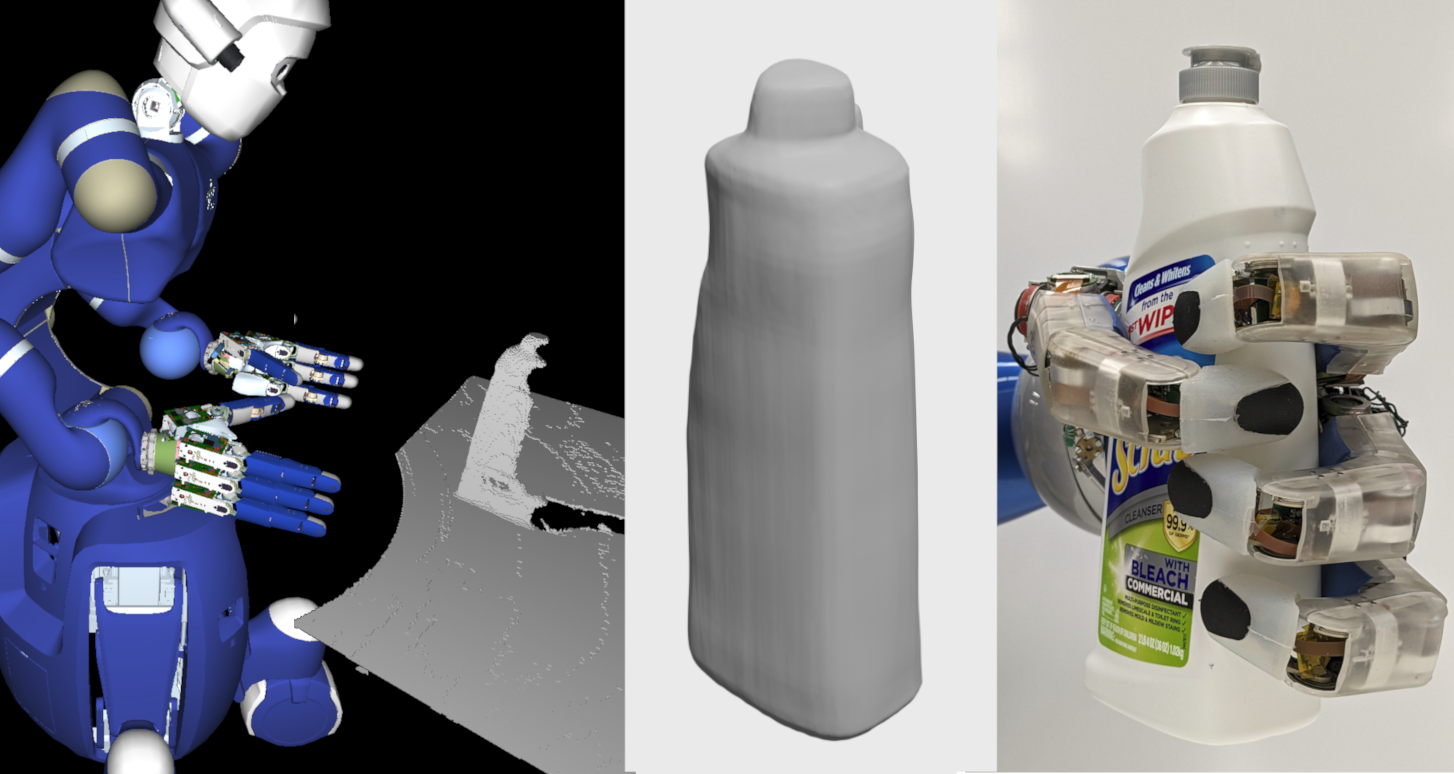}
\caption{Grasping the YCB bleach bottle using our grasping pipeline: The object is first perceived using Agile Justin's~\cite{Bauml2014} Kinect camera to obtain a single depth image. Afterward, the shape completion network predicts the full object shape, based on which the grasping network generates a stable grasp. The grasp is then executed on the real robot using whole-body motion planning ~\cite{Tenhumberg2022,Tenhumberg2023} for positioning the hand relative to the object and a kinematically calibrated robotic system~\cite{Tenhumberg2022a, Tenhumberg2023a}}
\vspace{-10pt}
\label{fig:hero}
\end{figure}

Using a dexterous multi-fingered robot hand allows a wide repertoire of complex grasps, leading to applications ranging from pick-and-place tasks~\cite{Wu2020} to object manipulation~\cite{Sievers2022} and to the utilization of tools via functional grasps~\cite{Wu2023}.
However, the exploitation of its full potential under the condition of very limited prior object knowledge requires a detailed reconstruction of the complete object shape and, based on this, a shape-specific prediction of the finger configuration.
Using these two critical components, we have integrated the first system that can predict and execute in real-time stable grasps with a fully actuated four-finger hand on a wide range of initially unknown object shapes.

Accurate and complete object shape reconstruction from partial and noisy depth data as can be acquired from a single viewpoint is challenging. We here take shape completion to the real world by introducing a procedure for synthetic training data generation and the training itself that can bridge the gap to the real sensing conditions.
Likewise, we use the same synthetic data generation procedure together with the trained shape completion network to acquire realistic object reconstructions for training the grasp network.
We further describe an extension to the grasp planner which increases the robustness of the generated training grasps against uncertainties in the relative pose between object and hand.

We show what is important in training shape completion and grasp prediction as a grasping pipeline for the real world.
Critical factors are the careful generation of training data as well as a training procedure that emphasizes difficult cases.

\begin{figure*}[t!]
    \centering
    \includegraphics[width=\linewidth]{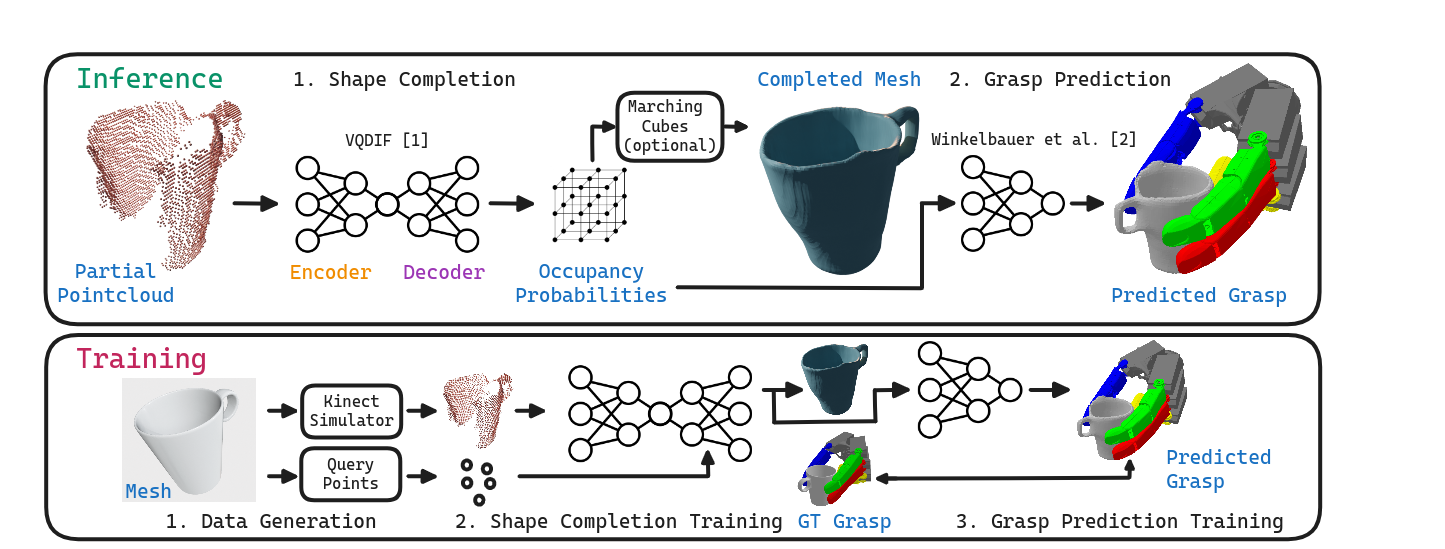}
    \caption{Our pipeline consists of two steps during inference. From a partial point cloud of an object obtained through rendering (training) or a depth sensor (inference), we use a shape completion network~\cite{Yan2022ShapeFormerTS} to predict the full object geometry implicitly as occupancy probabilities at query points on a grid from which we can optionally extract the completed mesh using Marching Cubes~\cite{Lorensen1987MarchingCA}. Once this network is trained, the predicted occupancy probabilities are used as input for the grasp predictor~\cite{Winkelbauer2022GraspPredictor} both during training on ground truth (GT) grasps as well as during deployment on our robot Agile Justin.}
    \vspace{-10pt}
    \label{fig:pipeline}
\end{figure*}

Our main contributions are:
\begin{itemize}
    \item A procedure for synthetic data generation and training of shape completion for improved sim2real transfer.
    \item A quantitative evaluation on synthetic and real data of the effects of the improved data and training procedure.
    \item An extension of the analytical grasp planner to generate grasps that are more resilient against uncertainties in the relative pose between hand and object.
    \item An adaptation of the grasp network architecture to handle ambiguities in the dataset without the requirement of generating multiple labels for each training sample.
    \item A validation of the proposed training procedures and of the system pipeline for reliable grasp generation and execution through a series of grasping experiments on our humanoid Agile Justin with a large range of initially unknown objects.
\end{itemize}

\section{Related Work}
Our work is related to deep learning based completion or reconstruction of partial 3D data. There is a large number of approaches to handle this modality from 3D convolutional neural networks~\cite{Wu20143DSA,Dai2016ShapeCU}, over point cloud-based methods~\cite{Yuan2018PCNPC,Xiang2021SnowflakeNetPC,Yu2021PoinTrDP} to 3D triangle mesh methods~\cite{Groueix2018AtlasNetAP,Liao2018DeepMC}.
In recent years, implicit function representations~\cite{Mescheder2018OccupancyNL,Peng2020ECCV,Chibane2020ImplicitFI,Xu2019DISNDI,Park2019DeepSDFLC,Yan2022ShapeFormerTS} have gained traction due to their strong performance on diverse problems.
While these works excel on benchmarks, they are seldom extended and evaluated under realistic conditions like varying viewpoints, object diversity and rotation, self-occlusion, and sensor artifacts. We make use of a network architecture introduced in~\cite{Yan2022ShapeFormerTS} but train it on a large, realistic dataset to enable generalization to real-world scenarios.

There is a strong connection to the robotic grasping literature, specifically works that rely on additional geometric understanding to perform grasps~\cite{Varley2016ShapeCE,Yan2017Learning6G,Merwe2019LearningC3,Chen2022ImprovingOG}. \citet{Varley2016ShapeCE} are the most similar to our work but only train on a limited number of objects (480 vs.\ 57,000) with a low-resolution voxel grid representation ($40^3$ vs. continuous implicit representation) and thus lack fidelity in completing unknown--and especially--real objects. Also, the grasps are optimized online using an analytical grasp planner, while we instantly predict stable grasps using our learning-based approach.
\citet{Yan2017Learning6G} only shows qualitative shape completion results and uses a parallel jaw gripper for grasping.
Also, \citet{Chen2022ImprovingOG} does not consider multi-fingered hands.
\citet{Merwe2019LearningC3} use an implicit function representation of the completed object and a multi-fingered hand but only show a single low-fidelity qualitative shape completion result on real data. Furthermore, the grasping prediction requires an online optimization, leading to long planning times.

In this work, we also propose a new method to handle ambiguities in the grasp training data.
This issue has been usually approached by generating multiple labels for training samples and then using the one closest to the network's prediction for the computation of the loss \cite{Winkelbauer2022GraspPredictor, Liu2019, liu2020deep}.
While this reduces the ambiguity problem, it makes the data generation process more computationally expensive.
In contrast to that, our proposed approach only requires one label per training sample and therefore does not suffer from that problem.

\section{Shape Completion}
We use implicit functions to represent the 3D surface of an object as the continuous decision boundary of a \textit{Deep Neural Network} (DNN) classifier predicting occupancy probability at each spatial location.
Compared to discrete representations, implicit functions have the advantages of infinite resolution at a low memory footprint and are agnostic to the genus of the object geometry.
During inference, the implicit function can be queried densely on a voxel grid and meshed using \textit{Marching Cubes}~\cite{Lorensen1987MarchingCA} to extract an explicit mesh representation of the object.
These networks are trained by giving partial observations like depth images or point clouds as input and supervising with occupancy labels at sampled spatial locations.

Specifically, we use the VQDIF~\cite{Yan2022ShapeFormerTS} (Vector Quantized Deep Implicit Functions) network architecture which improves on Convolutional Occupancy Networks~\cite{Mescheder2018OccupancyNL,Peng2020ECCV} using a more compact sequence-based shape representation.

\subsection{Training data generation}
We leverage the ShapeNetCore (v1) subset of the ShapeNet~\cite{Chang2015ShapeNetAI} dataset containing more than 57,000 3D models across different categories to generate training data.

Many of the provided meshes are, however, not watertight.
We, therefore, first pre-process the entire dataset using the mesh-fusion~\cite{Stutz2018ARXIV} pipeline with optimizations for speed and parallel execution to allow for the processing of large-scale datasets. We apply quadratic edge collapse decimation~\cite{MeshLab} and small disconnected component removal to reduce the triangle count by 95\% and remove unwanted mesh-fusion artifacts.
We sample 100,000 points from the surface of the watertight meshes.
For supervision of the occupied space, we use multiple sampling strategies.
We uniformly sample 100,000 points in the unit bounding cube of each object with a small padding of 0.1 on each side and determine their occupancy value using a triangle intersection test as in~\citet{Mescheder2018OccupancyNL}.
We further sample 10,000 points from the surface of the object and add Gaussian noise with various standard deviations for a total of an additional 100,000 points to increase the resolution of the supervision signal close to the surface similar to~\citet{Xu2019DISNDI}.
Additionally, we sample 100,000 points from spheres centered on the object with various radii to prevent missing occupancy information inside the unit cube under object rotation.
We refer to these points as \textit{query points} in the remainder of this article.

\begin{figure}[htbp]
    \centering
    \vspace{-10pt}
    \includegraphics[width=\linewidth]{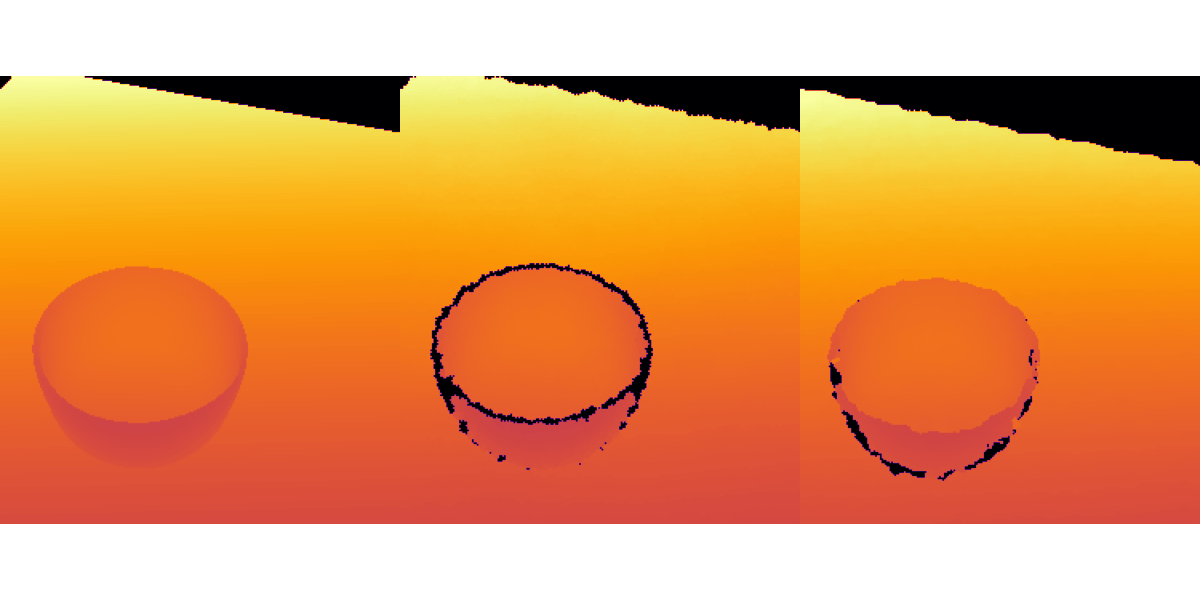}
    \vspace{-20pt}
    \caption{Comparison of standard depth rendering (left), Kinect depth simulation (middle), and the real Kinect depth image (right). The object mesh is obtained via a laser scanner, and its pose through a registration step.}
    \label{fig:depth_comp}
\end{figure}

To simulate realistic partial views, we render depth, surface normal, and simulated Kinect depth images from views distributed on the upper hemisphere facing each object using an optimized version of the \textit{Blensor}~\cite{Gschwandtner2011BlenSorBS,Bohg2014RobotAP} Kinect sensor model. Figure \ref{fig:depth_comp} shows a comparison of a scene rendered using a standard renderer, the Kinect simulator and the real Kinect sensor data. The simulation more closely resembles the real depth image, showing similar noise patterns and regions of missing data compared to the standard depth rendering. At the same time, there are still considerable differences and room for improvement to more accurately simulate real sensor artifacts.

For each object, we render 100 views with the camera positioned at varying distances for a total of around 5.7 million views which we split 9 to 1 per object category into training and validation sets. A test set is not necessary as we perform testing on novel objects and real data, as explained later.
We also generate additional data where we randomly scale each object along its upright axis by up to 20\% for every rendered view to further increase the object diversity, especially for categories with few examples.

The complete data generation pipeline is summarized in algorithm \ref{alg:data}.
\begin{algorithm}
\caption{Data Generation}
\begin{algorithmic}[1]
\STATE Obtain ShapeNetCore subset containing $N$ meshes
\FOR{$i=1$ \TO $N$}
\STATE Make mesh watertight
\STATE Normalize mesh
\STATE Sample 100,000 points from mesh surface
\STATE Sample 100,000 random points in unit cube
\FOR{$r=1$ \TO $5$}
\STATE Sample 100,000 random sphere points (radius $r$)
\ENDFOR
\FOR{$s$ \textbf{in} $S\in\{0.001,0.25\}$}
\STATE Sample 10,000 points from the mesh surface
\STATE Sample and add noise$~\mathcal{N}(0,s)$
\ENDFOR
\STATE Compute occupancy for all points
\FOR{$v=1$ \TO $100$}
\STATE Apply random scale to mesh (optional)
\STATE Sample view on upper hemisphere
\STATE Render depth, normal and Kinect depth images
\ENDFOR
\ENDFOR
\STATE Split data into 90\% training and 10\% validation sets
\end{algorithmic}
\label{alg:data}
\end{algorithm}

\subsection{Training}
\label{sec:training}
We train our shape completion networks in a fully supervised manner.
For each training example, we randomly sample one of the ShapeNet objects and then one of its corresponding rendered depth or simulated Kinect depth images. We refer to the DNN trained on the simulated Kinect depth images as \textsc{kinect} and the one trained on data with randomly scaled objects during data generation as \textsc{kinect scale}.
The depth image is projected to a 3D point cloud based on the known camera intrinsic parameters used during rendering. For the \textsc{basic} training setup, we add Gaussian noise to all points and additional noise to edge points which we partially remove as in~\citet{humt2023shape}.
We center the input data at its origin and scale its longest side to unit length.

We load the uniform, noisy surface and sphere query points for the given object, crop points outside the unit bounding cube, subsample 2048 queries, and transform them to match the scale and camera frame of the normalized input data.
Since the robot's base frame of reference is known during deployment, we further transform all points from camera coordinates to the robot's world coordinate system using the robot's kinematic chain.
This reduces the complexity of the learning problem by removing camera tilt.
At each query point the occupancy probability is predicted and the ground truth binary labels are then used to compute the binary cross-entropy loss, which we use as training signal.

Once the training is complete, we improve the model performance further through an automated finetuning step. Akin to \textit{Hard Negative Mining}, we employ an importance sampling strategy by oversampling object views proportional to the magnitude of their obtained loss. To mitigate the high class imbalance of ShapeNet (there are approximately 35 times more cars than mugs), we further resort to weighted batch sampling using the inverse class frequency as weights.
The effects of this additional step are exemplified in Figure \ref{fig:finetuning}.
%

All methods are implemented in the \textit{C++} and \textit{Python} programming languages. We use \textit{PyTorch}~\cite{Paszke2019PyTorchAI} and \textit{Lightning}~\cite{PyTorchLightning2019} as our primary Deep Learning framework. Networks are Trained with the \textit{AdamW}~\cite{Kingma2014AdamAM} optimizer with a learning rate of $1e-4$ and weight decay of ${0.01}$. All networks are trained for a maximum of 100 epochs with learning rate reduction when performance on the validation set plateaus for 10 epochs and early stopping after 30 epochs without improvement.

\section{Grasp prediction}
\label{sec:grasping}

To efficiently predict grasps for a given unknown object, we are building on top of the learning-based approach proposed by~\citet{Winkelbauer2022GraspPredictor}.
Here, grasping is approached as a supervised learning problem, learning the mapping from a given observed object $o$ to a set of grasps $\{x_1, ..., x_n\}$. Each grasp $x_i$ is composed of a 6D hand pose $h$ and a joint configuration $q \in \mathbb{R}^{12}$. 
The ground truth grasp annotations for synthetic training objects are generated using an analytical grasp planner.
The grasp planner formulates grasping as an optimization problem by using a grasp quality metric as objective.
Our quality metric is based on the epsilon quality metric \cite{Ferrari1992}, which is defined as the magnitude of the minimal external wrench that would break the grasp.
As learning architecture, a two-stage neural network is used.
The first stage is a generative network for hand poses that learns the mapping $\mathcal{G}: o \mapsto \text{p}(h|o)$.
The second stage is a regressive network which learns the mapping $\mathcal{J}: (o, h) \mapsto (q, \Delta h, s)$, predicting a joint configuration $q$, a small hand pose correction $\Delta h$, and a grasp quality $s$ for each generated hand pose $h$.
For further details, we refer the reader to \citet{Winkelbauer2022GraspPredictor}.

\subsection{Object pose uncertainty}
\label{sec:pose_uncertainty}

\begin{figure}[htbp]
\centering
\subfloat[All contact points]{%
  \centering%
  \includegraphics[width=0.5\linewidth]{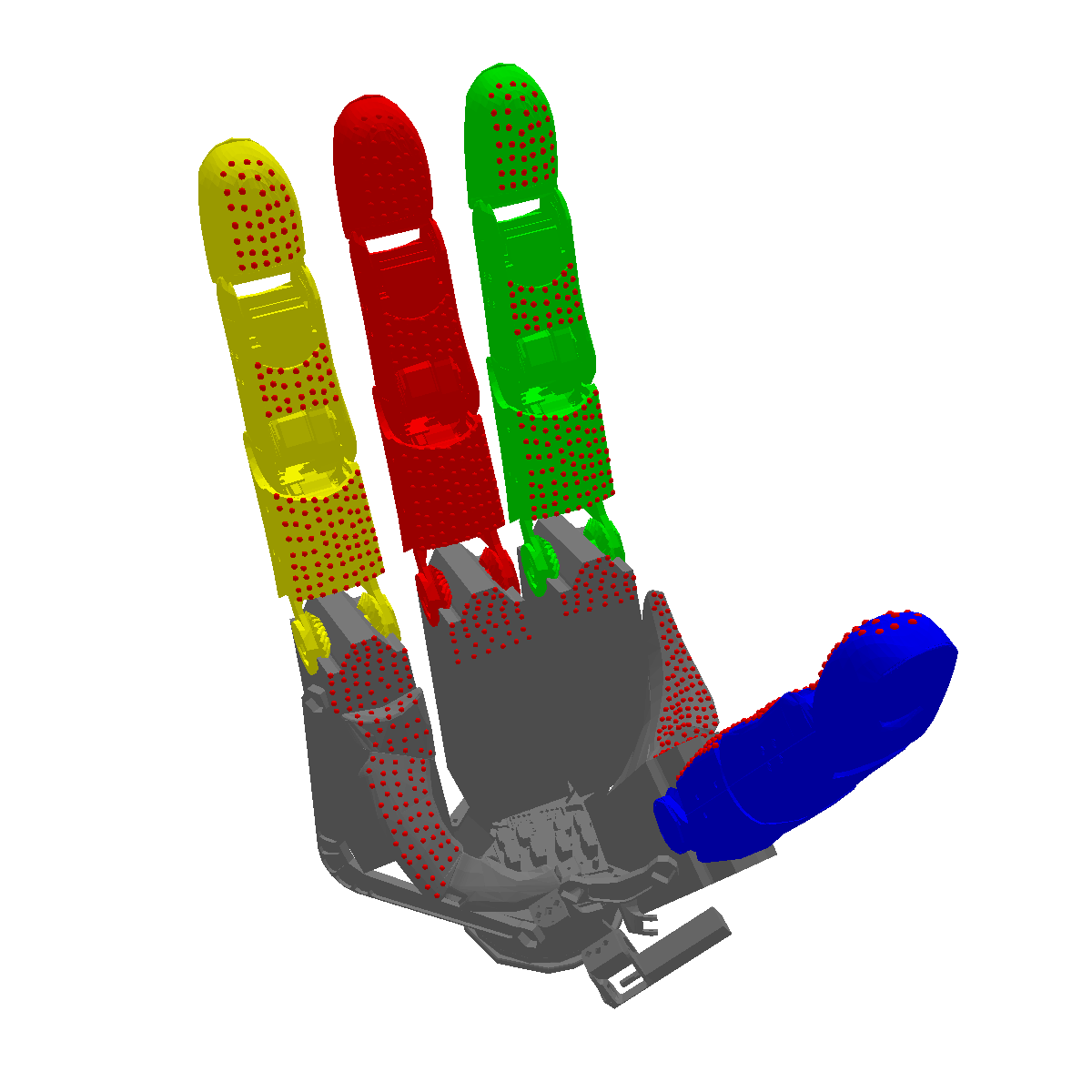}%
}
\subfloat[Central contact points]{%
  \centering%
  \includegraphics[width=0.5\linewidth]{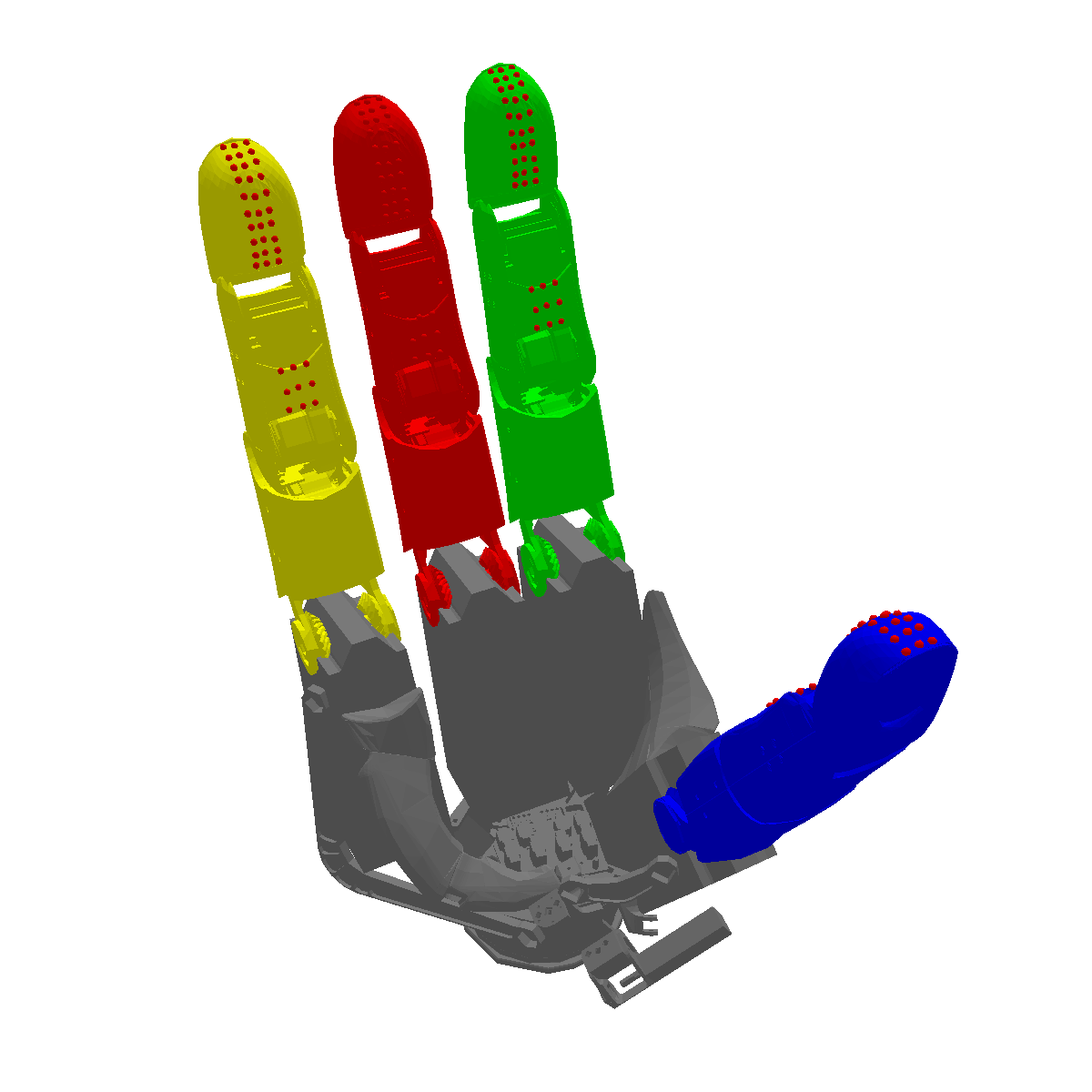}%
}
\caption{The DLR-Hand II~\cite{Butterfass2001} which we use in our experiments. Each red dot represents a potential contact point. In the left image, all contact points used in the grasp planner are shown, while the right image shows only the central contact points that are used to verify that a finger is still in full contact with the object after slight pose variations.}
\vspace{-10pt}
\label{fig:hand}
\end{figure}

When executing a grasp on the real system, multiple small inaccuracies in the grasping pipeline lead to an uncertainty in the actual relative positioning of object and hand.
These inaccuracies can occur in the calibration of the camera and kinematic structure, the shape completion, and also the grasp prediction itself.
Some grasps, however, require a very precise placement of the fingers to be successful.
So, due to small inaccuracies, a finger might miss or slide over an edge, which can lead to a distortion of the grasp and even to dropping the object.
To prevent such grasps, we adapt our grasp planner's objective in a way that we diminish the quality of grasps that are prone to missing the object due to small changes in its relative pose.
To detect such grasps, we check the grasp under multiple small object pose deviations.
In detail, we move the object pose using a translation $\Delta t_i$.
To then resolve possible collisions with the hand base, the hand is moved backward along the approach direction until any collision with the hand base is resolved.
Afterward, the fingers are closed until they touch the object or until they reach the limits of the joints.
For each joint, we check whether the difference between the resulting joint configuration and the original one is smaller than \SI{23}{\degree}.
We further check whether each finger, that was in contact with the object before, still contacts the object in at least one contact point close to the center of the finger (see \cref{fig:hand}).
This detects situations where a finger is still in contact with the object but only touches its surface slightly with the side of the finger.
If one of the checks fails, the grasp quality is decreased to \SI{30}{\percent} of its original computed value.
To keep the grasp planner's objective computationally efficient and deterministic, we test six translations $\Delta t_i$, each moving the object along one axis for \SI{2}{\centi\meter} in both directions.
In summary, grasps that are unstable when there is object pose uncertainty are devalued and therefore, such grasps are prevented by the grasp planner, if possible.

\subsection{Ambiguities in predicting joint configurations}
\label{sec:ambiguity}

\begin{figure}[htbp]
\centering
\resizebox{0.8\linewidth}{!}{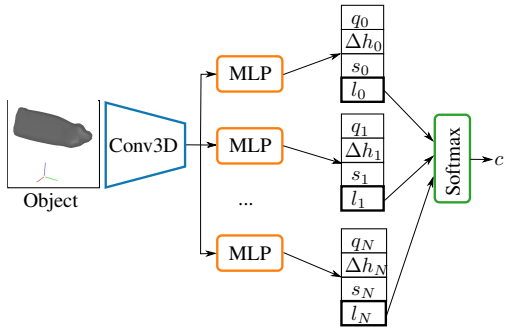}%
\caption{Adapted joint predictor network architecture with multiple heads. Each head predicts one configuration together with a logit $l_k$, which is used to determine which head to use at inference.}
\vspace{-10pt}
\label{fig:ambiguity_net}
\end{figure}

As mentioned in the introduction of \cref{sec:grasping}, the second stage of the grasp prediction network, is trained to predict for a given object $o$ and hand pose $h$ the joint configuration $q$ with the highest resulting grasp quality.
This mapping is ambiguous, however, meaning there are usually multiple valid joint configurations that lead to the same or very similar grasp qualities.
When using standard supervised training, this leads to learning the mean of all valid joint configurations, which, however, might not be valid.
The way this is handled by \citet{Winkelbauer2022GraspPredictor}, is by explicitly having multiple ground truth annotations for each training sample and then using the one closest to the network prediction for computing the loss.
The disadvantage of this approach is that it is computationally expensive to find this set of joint configurations.
Also, this might lead to including joint configurations with less quality and consequentially to the network learning not the top grasp.

We here present an alternative solution to the ambiguity problem, which is based on adapting the network architecture as shown in Figure \ref{fig:ambiguity_net}.
Instead of predicting just one configuration, we let the network predict $N$ configurations using $N$ heads.
Each head $k$ predicts a joint configuration $q_k$, a small hand pose correction $\Delta h_k$, and a grasp quality $s_k$.
To compute the loss, we only use the prediction of the head $j$ that is closest to the ground truth $\widehat{q}$: $j = \argmin_{k} \ltwo{\widehat{q} - q_k}$.
The joint loss term is hence defined as

\begin{equation}
L_\text{joint} = \ltwo{\widehat{q} - q_j}
\,.
\end{equation}
The loss terms for the predicted grasp quality $s_j$ and hand pose correction $\Delta h_j$ are formulated in an analog manner.
To decide which head's prediction to use at inference time, we additionally model the relative frequency of each head's prediction in the training dataset.
In detail, each head $k$ additionally predicts a logit $l_k$, which are together transformed into a probability distribution using softmax: $c_k = \frac {e^{l_k}}{\sum _{j=1}^{K}e^{l_j}}$
Using the cross entropy loss function, this distribution is optimized to model the true underlying distribution of the predicted modes in the training dataset:

\begin{equation}
L_\text{class} = - \log{c_j}
\end{equation}
At inference time, we select the prediction of the head $k$ with the largest $l_k$, which corresponds to taking the prediction that is the most prominent in the training dataset.
Using this architectural adaptation, it is possible for the network to handle ambiguities in the dataset, while only one ground truth label for each training sample is necessary. 
In this work, we are using $N = 5$ heads, which we found to be enough to cover the main ambiguities contained in our training data.

\section{Experiments}
\label{sec:experiments}
We conduct a systematic evaluation of our shape completion pipeline through multiple experiments on synthetic and real test data. First, we quantitatively assess the generalization capabilities to novel objects using volumetric and surface metrics on the completed shapes from simulated depth images. Next, we validate performance on real partial scans from a Kinect sensor and evaluate the ability to transfer from synthetic training data. Finally, we provide a qualitative grasp analysis by attempting robotic grasping of the completed shapes from real partial views. These experiments analyze the utility of our approach for enabling robotic manipulation of unknown objects based on incomplete observations from commodity RGB-D sensors. The robust performance demonstrates accurate shape inference on synthetic and real partial inputs, along with successful grasping by leveraging the completed geometry.

\subsection{Shape completion}
\label{sec:exp_shape_completion}
The performance of shape completion is evaluated on both synthetic and real test data. First, we assess the ability to generalize to novel objects not seen during training. We use 29 objects which were selected and scanned with a laser scanner for an Automatica trade fair demonstration as well as 38 objects of the YCB~\cite{YCB} dataset, for which we render 100 simulated Kinect depth images per object, each covering different viewpoints. We quantitatively evaluate the completed shapes using volumetric metrics like intersection over union (IoU), F1-score, precision, and recall computed by sampling one million spatial points uniformly inside the unit cube (Table \ref{tab:exp_synth_vol}). Additionally, we extract surface meshes for each completion and measure point-based versions of F1, precision, recall, as well as Chamfer (L1) distance~\cite{Mescheder2018OccupancyNL,Tatarchenko2019WhatDS} between points sampled from the surface of the ground truth and generated meshes (\ref{tab:exp_synth_mesh}).

\begin{table}[htbp]
\caption{Volumetric results on Automatica/YCB dataset}
\centering
\begin{tabular}{lrrrr}
\toprule
\textbf{Model} & \textbf{IoU}$\uparrow$ & \textbf{F1}$\uparrow$ & \textbf{Precision}$\uparrow$ & \textbf{Recall}$\uparrow$ \\
\midrule
 \textsc{kinect finetune} &           66.7 &                75.7 &                 73.5 &              83.3 \\
 \textsc{kinect scale}    &           60.4 &                71   &                 75.5 &              73.6 \\
 \textsc{kinect}          &           58   &                68.9 &                 75.7 &              70.6 \\
 \textsc{basic}           &           49.5 &                61.1 &                 74   &              59.5 \\
\bottomrule
\end{tabular}
\label{tab:exp_synth_vol}
\end{table}

Table \ref{tab:exp_synth_vol} shows the volumetric evaluation results on the Automatica/YCB dataset for different variants of our shape completion model. We observe that fine-tuning the network (\textsc{kinect finetune}) as explained in section \ref{sec:training} leads to the best performance with an IoU of 66.7\% and F1-score of 75.7\%. The network trained on Kinect data but without fine-tuning (\textsc{kinect}) performs worse, justifying the additional step. Interestingly, the network trained on simulated Kinect data with additional random scaling augmentation (\textsc{kinect scale}) leads to considerable improvement. Finally, the network trained only on synthetic data without simulation (\textsc{basic}) performs significantly worse, with an IoU of just 49.5\%, highlighting the domain gap between synthetic and real data. Overall, the results validate our design choices of using Kinect simulation, scaling augmentation, and finetuning, which enable accurate shape completion from real partial scans.

\begin{table}[htbp]
\caption{Mesh surface results on Automatica/YCB dataset. Chamfer-L1 (CD)$\times10$.}
\centering
\begin{tabular}{lrrrr}
\toprule
\textbf{Model}    &   \textbf{CD}$\downarrow$ &   \textbf{F1}$\uparrow$ &   \textbf{Precision}$\uparrow$ &   \textbf{Recall}$\uparrow$ \\
\midrule
 \textsc{kinect finetune} &                            0.258 &                          41.5 &                           43.4 &                        40.2 \\
 \textsc{kinect scale}    &                            0.276 &                          43.4 &                           44.6 &                        42.8 \\
 \textsc{kinect}          &                            0.29  &                          42.7 &                           44.1 &                        42   \\
 \textsc{basic}           &                            0.455 &                          34.1 &                           36.8 &                        32.6 \\
\bottomrule
\end{tabular}
\label{tab:exp_synth_mesh}
\end{table}

Table \ref{tab:exp_synth_mesh} shows the mesh-based metrics on the Automatica/YCB dataset. We again observe the benefit of Kinect pretraining and finetuning, with \textsc{kinect finetune} achieving the lowest Chamfer-L1 distance of 0.258. The network trained only on synthetic data (\textsc{basic}) has significantly higher CD, revealing the domain gap.
Overall, the mesh metrics confirm that leveraging simulated Kinect data along with finetuning is crucial for accurate shape completion from real partial observations, enabling tasks like robotic grasping.

\begin{figure}[htbp]
\centering
\includegraphics[width=\linewidth]{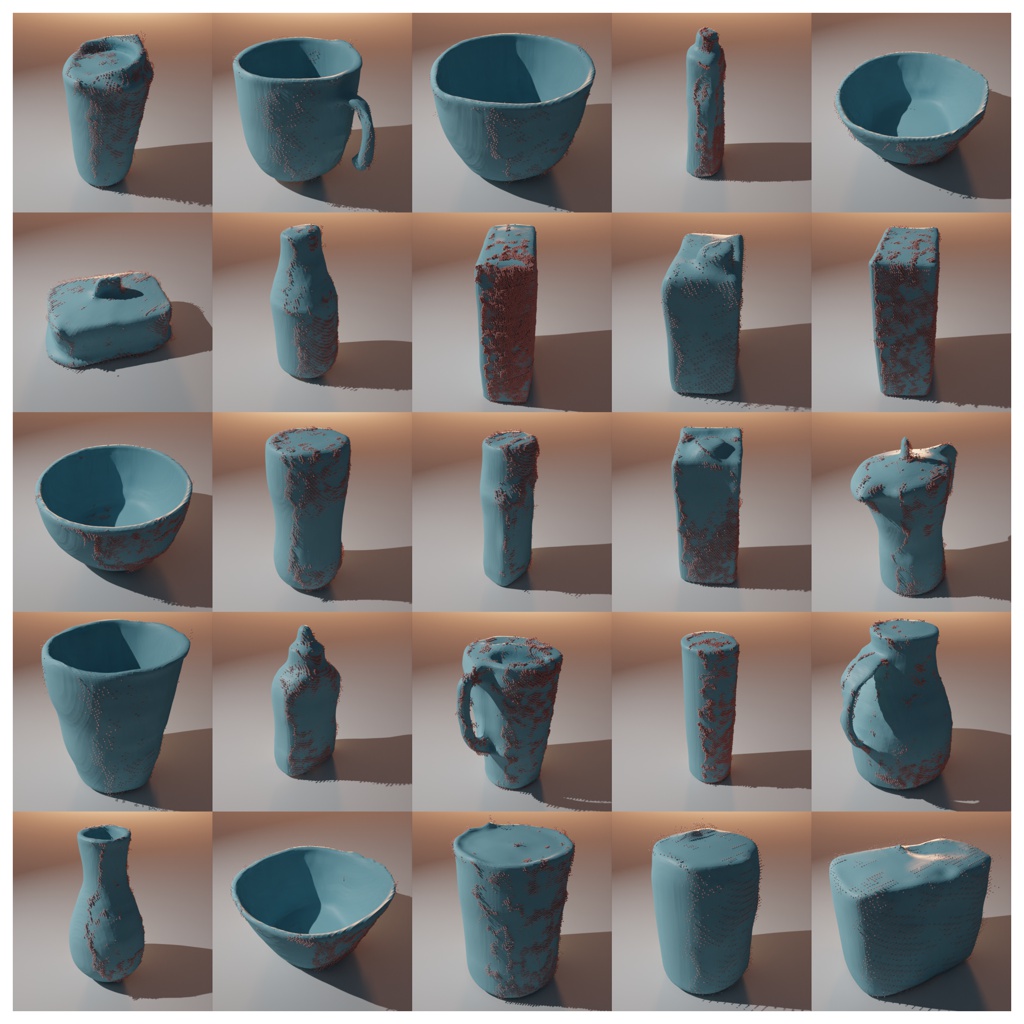}
\caption{Qualitative results on the Automatica/YCB dataset. Our approach yields detailed object geometry from partial, noisy inputs. The complete meshes correspond to our \textsc{kinect finetune} model and its quantitative results in table \ref{tab:exp_real}.}
\vspace{-10pt}
\label{fig:exp_real_qualitative}
\end{figure}

Second, we validate performance on real sensor data from a Kinect camera (Table \ref{tab:exp_real} and Figure \ref{fig:exp_real_qualitative}). We use a limited subset of 25 objects and a single view.
To align the ground truth meshes with the real data, we use a global-to-local, coarse-to-fine pose estimation pipeline\footnote{\url{https://github.com/hummat/easy-o3d}} with manual visual inspection of the result. Note that this is only necessary for quantitative evaluation of the results and \textit{not} during the inference phase when deploying the pipeline on the robot.

We separate the object points from the complete point cloud obtained from the depth image projection through plane segmentation. We first fit a plane into the scene corresponding to the table surface and then only keep points that are above. Due to noise, this procedure removes points close to the surface, like the bottom of bowls. To prevent this, we compute the convex hull of the object points, slightly increase its size, and translate it towards the table. We then use it as a mask to cut points from the table surface and reintroduce them to the object points.
The reconstructed shapes are evaluated using the same set of volumetric and surface metrics to quantify the sim-to-real transferability.

\begin{table}[htbp]
\caption{Results on real scans from Automatica/YCB objects. Chamfer-L1 (CD)$\times10$.}
\centering
\begin{tabular}{lrrrrrrrr}
\toprule
\textbf{Model} & \textbf{CD}$\downarrow$ & \textbf{IoU}$\uparrow$ & \textbf{F1}$\uparrow$ &   \textbf{Precision}$\uparrow$ & \textbf{Recall}$\uparrow$ \\
\midrule
\textsc{kinect finetune}      &       0.134 &     74.7 & 83   &        79.2 &     89.2 \\
\textsc{kinect scale}         &       0.147 &     72   & 81.1 &        79.8 &     84.2 \\
\textsc{kinect}               &       0.128 &     74.2 & 82.5 &        81.3 &     85.3 \\
\textsc{basic}                &       0.143 &     71.2 & 80.4 &        82.3 &     80.7 \\
\bottomrule
\end{tabular}
\label{tab:exp_real}
\end{table}

\begin{figure}[htbp]
\centering
\subfloat[Basic depth rendering]{%
  \centering%
  \includegraphics[width=0.4\linewidth]{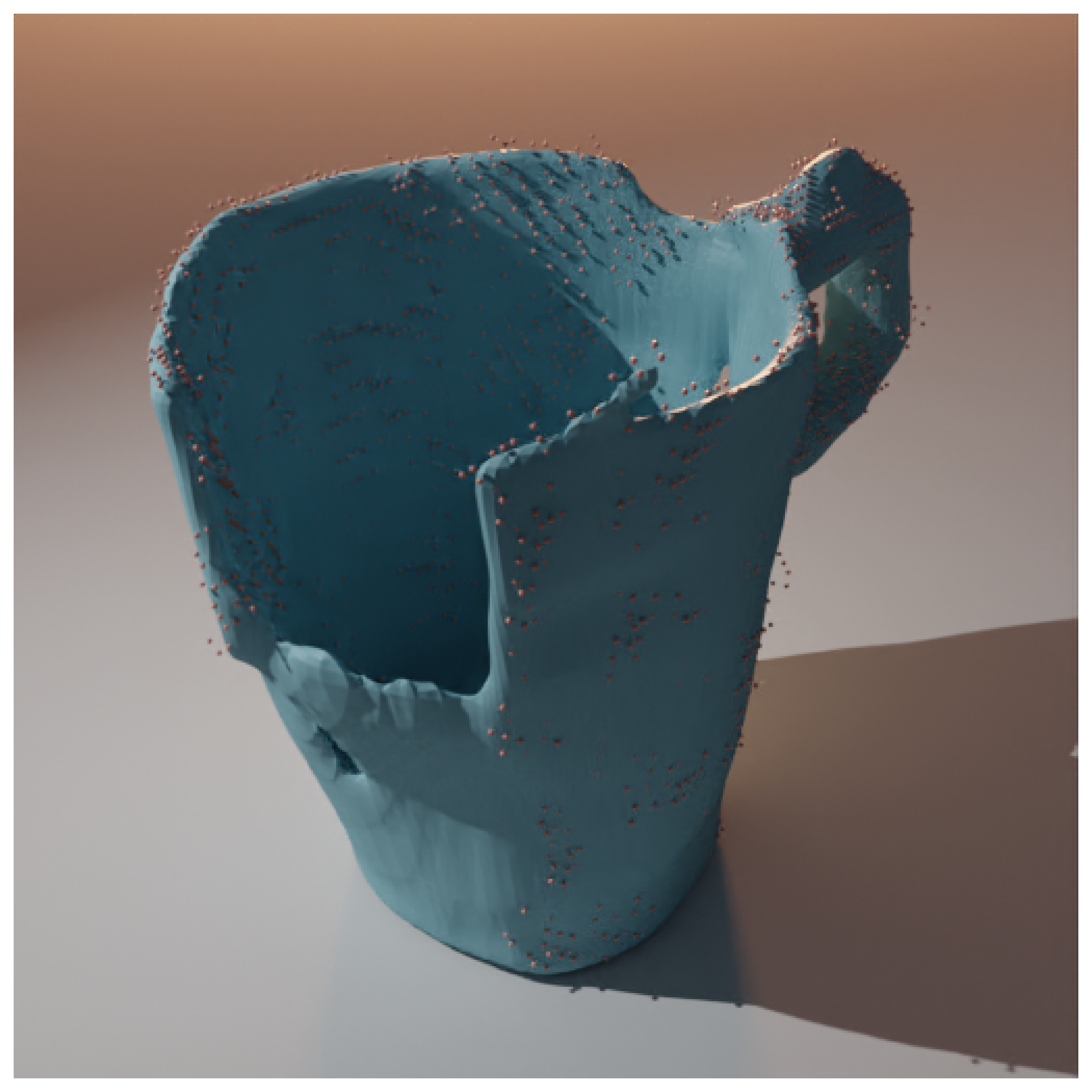}%
  \label{fig:standard}%
}
\subfloat[With Kinect simulation]{%
  \centering%
  \includegraphics[width=0.4\linewidth]{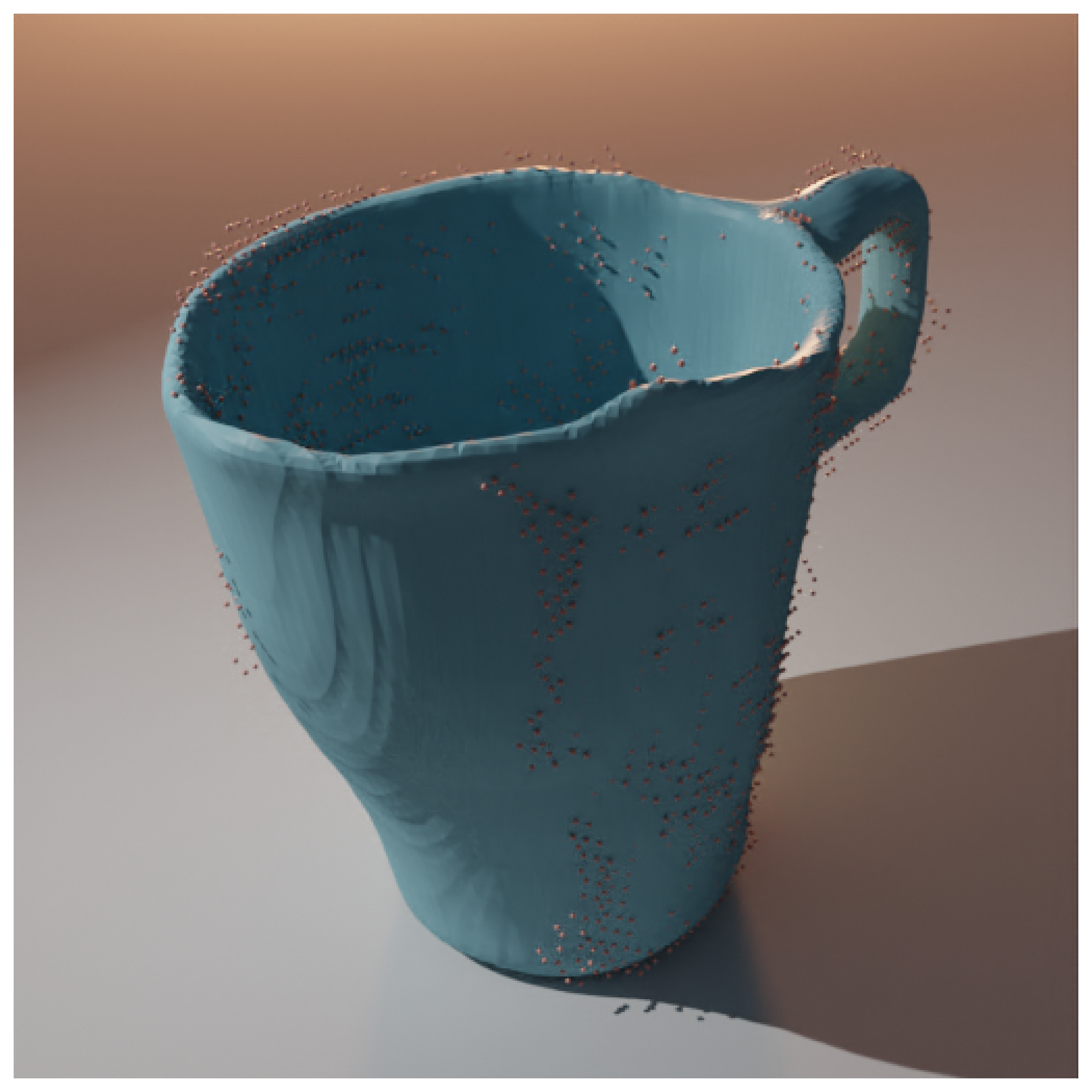}%
  \label{fig:simulation}%
}
\caption{Effects of Kinect depth simulation. Training on unrealistic depth data leads to failure in challenging regions far away from input data and training distribution (left). Accurately simulating sensor characteristics resolves this (right).}
\vspace{-15pt}
\label{fig:kinect}
\end{figure}

\begin{figure}[htbp]
\centering
\subfloat[No finetuning]{%
  \centering%
  \includegraphics[width=0.4\linewidth]{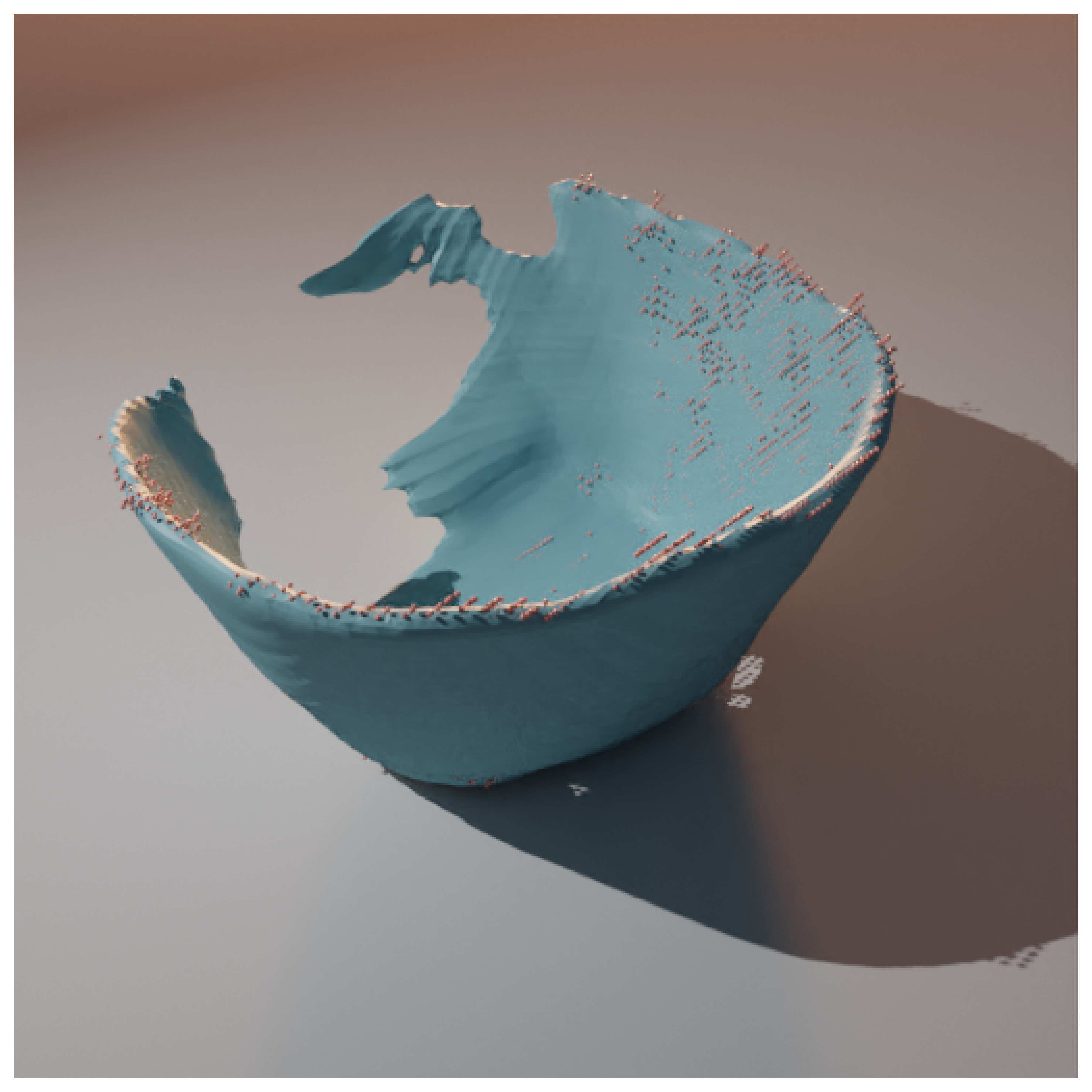}%
  \label{fig:no_finetuning}%
}
\subfloat[With finetuning]{%
  \centering%
  \includegraphics[width=0.4\linewidth]{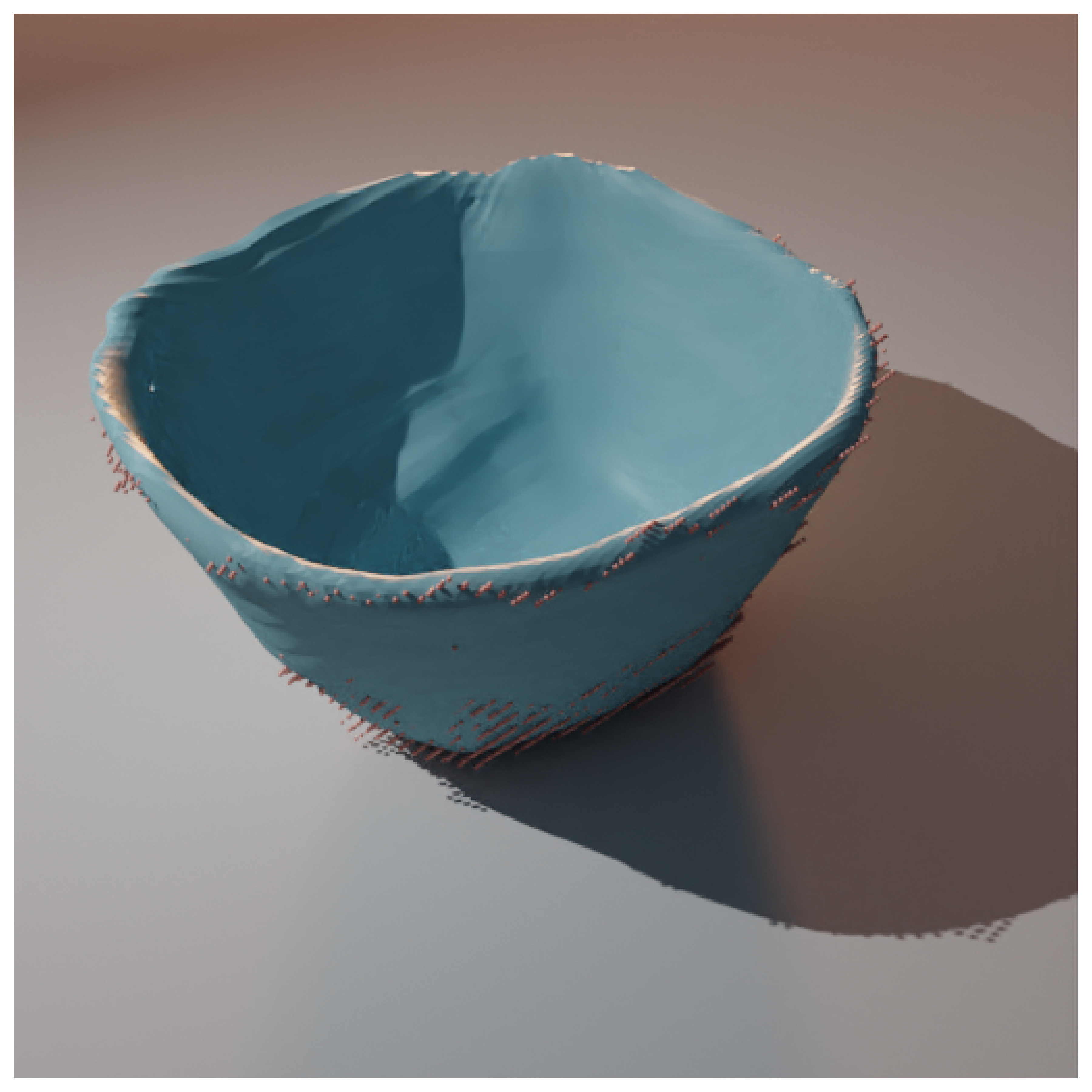}%
  \label{fig:with_finetuning}%
}
\caption{Effects of finetuning. The standard training methodology can lead to failure cases on challenging objects and viewpoints. Automated finetuning on difficult examples through importance sampling alleviates or greatly reduces this problem.}
\vspace{-10pt}
\label{fig:finetuning}
\end{figure}

Table \ref{tab:exp_real} shows the evaluation on real test scans from Automatica/YCB objects. We observe that the models on simulated Kinect data achieve lower Chamfer distances and higher volumetric scores compared to the basic training, consolidating the results from the synthetic data evaluation and the qualitative results, showing failure cases without Kinect simulation (Figure \ref{fig:kinect}) and without finetuning (Figure \ref{fig:finetuning}) and high-fidelity completions for a diverse set of objects and viewpoints (Figure \ref{fig:exp_real_qualitative}).
The \textsc{kinect} model performs best in terms of Chamfer distance but worse than the finetuned model on the volumetric metrics. The scale-augmented Kinect model performs worst in this setting. Due to the small sample size and the resulting high variance--all methods show a standard deviation of around $0.05$ for Chamfer distance and $2.4\%$ (absolute) for IoU--further investigation is needed to differentiate between the different Kinect methods.

\subsection{Grasping}

Using the adapted grasp planner described in \cref{sec:pose_uncertainty}, we generate a dataset of \num{190000} grasps across \num{12000} ShapeNet objects~\cite{Chang2015ShapeNetAI}.
For the training of the second stage, we generate an additional \num{100000} grasps using hand poses generated by the trained first stage.
This makes sure the second stage can correctly process the hand poses produced by the generative network.
For each ShapeNet object in the training dataset, we generate a simulated Kinect depth image, based on which the full shape is again reconstructed using the shape completion network \textsc{kinect finetune}, presented in the last section.
The two-stage grasp network, including the improvements described in \cref{sec:ambiguity}, is now trained using the generated shape completions as input and the optimized stable grasps as labels.

We evaluate the grasping stage in simulation on the same dataset as used in \cref{sec:exp_shape_completion}.
Each object is grasped 10 times, each time rotated randomly around the up-axis.
Of all tested \num{660} grasps, \SI{95.2}{\percent} were successful.	
The most prominent failure case is the object being too small to be grasped while keeping a distance to the table.
Same as with the shape completion network, all evaluated objects shown here or in the video accompanying the paper are unknown to the grasping network and not part of the training dataset.

The full pipeline of predicting \num{1000}~grasps for a given object takes only about \SI{1}{\second} with shape completion \SI{0.7}{\second} and grasp prediction \SI{0.3}{\second}~\footnote{Intel Xeon 6144 with 16 cores and a single GPU NVIDIA V100.}.

As visible in the video, even for unusual objects (e.g., the stuffed dog), the completion only gradually breaks down, leading to locally correct completions that still allow for a successful grasp.

In the following subsections, the two adaptations proposed to the grasp network are examined more closely:

\subsubsection{Object pose uncertainty}

\begin{figure}[htbp]
\centering
\subfloat[Without pose uncertainty]{%
  \centering%
  \includegraphics[width=0.5\linewidth]{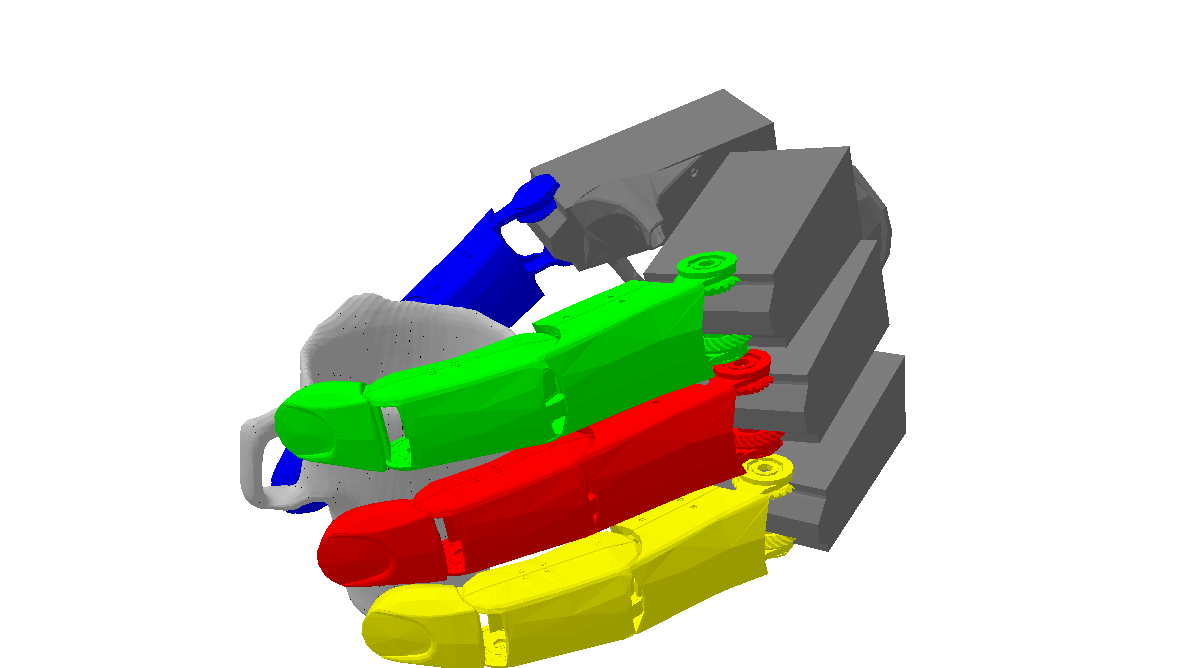}%
  \label{fig:uncertainty_cup_old}%
}
\subfloat[With pose uncertainty]{%
  \centering%
  \includegraphics[width=0.5\linewidth]{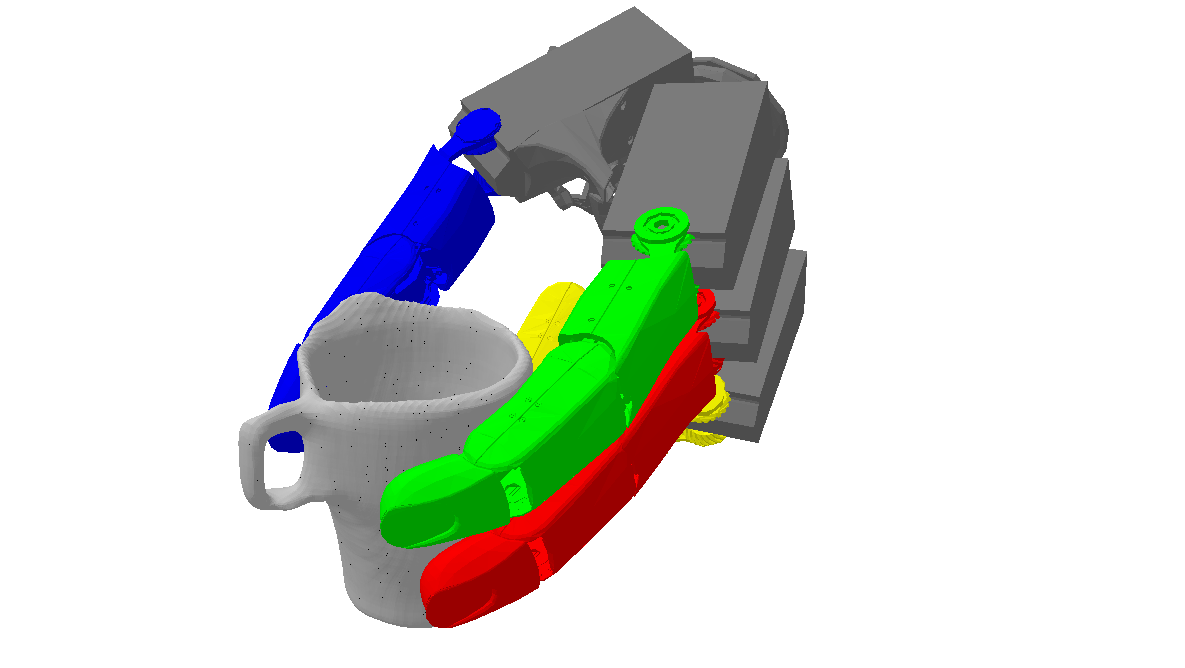}%
  \label{fig:uncertainty_cup_new}%
}
\caption{Comparing grasp predictions with and without considering object pose uncertainty during training data generation: When taking pose uncertainty into account, it leads to fingers being placed further away from edges.}
\label{fig:exp_pose_uncertainty}
\vspace{-10pt}
\end{figure}

When not taking the object pose uncertainty into account, fingers are sometimes placed close to or even on object edges, as they usually increase the grasp quality.
However, when performing the grasp on the real system, such placed fingers might slide over the edge and involuntary rotate the object or even break the grasp
Using the changes proposed in \cref{sec:pose_uncertainty}, the grasps in our training dataset and, therefore, also the grasps predicted by the network, are chosen in a way that fingers are placed further from edges, if possible.
This can be seen in \cref{fig:exp_pose_uncertainty}. The network trained without considering object pose uncertainty is trying to fit as many fingers as possible on the small surface of the cup, leading to one finger being placed on the edge of the cup
The network trained with object pose uncertainty, however, instead only uses three fingers, which allows them to be placed at sufficient distance to the object edges.
So even with slight variations in the object pose, the grasp would stay stable.

\subsubsection{Joint configuration ambiguity}

\begin{figure*}[htbp]
\captionsetup[subfigure]{labelformat=empty}
\centering
\subfloat[$s = 7.85$]{%
  \centering%
  \includegraphics[width=.14\linewidth]{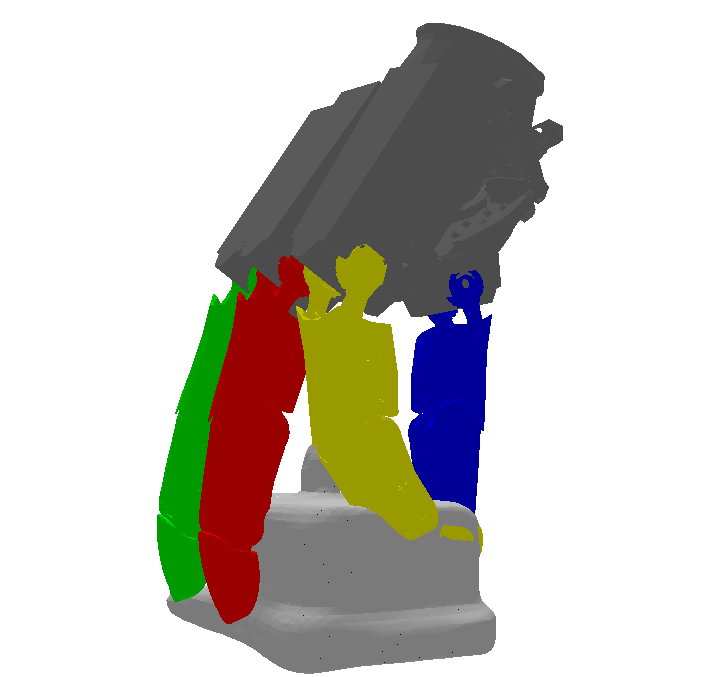} %
}
\unskip\ \vrule\
\subfloat[$s_0 = 1.37$ \\ $c_0 = \SI{3}{\percent}$]{%
  \centering%
  \includegraphics[width=.14\linewidth]{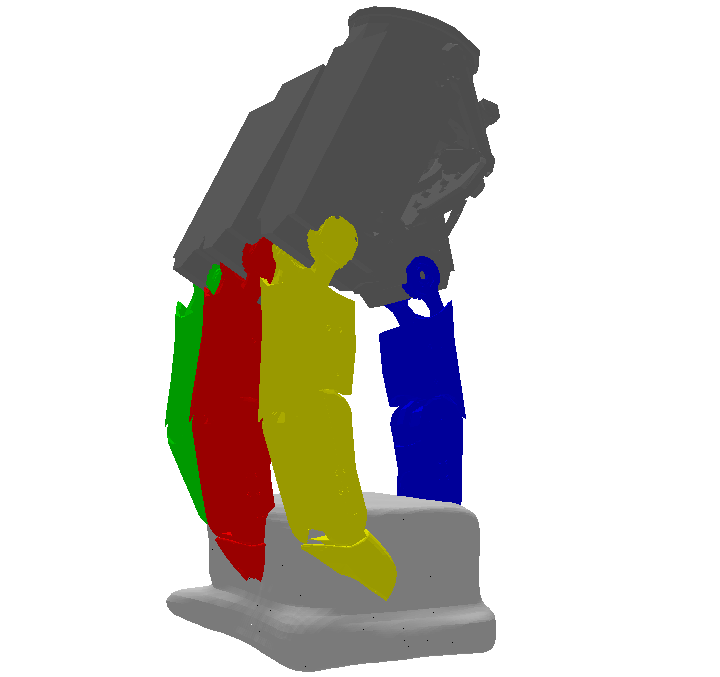} %
}
\subfloat[$s_1 = 8.93$ \\ $c_1 = \SI{50}{\percent}$]{%
  \centering%
  \includegraphics[width=.14\linewidth]{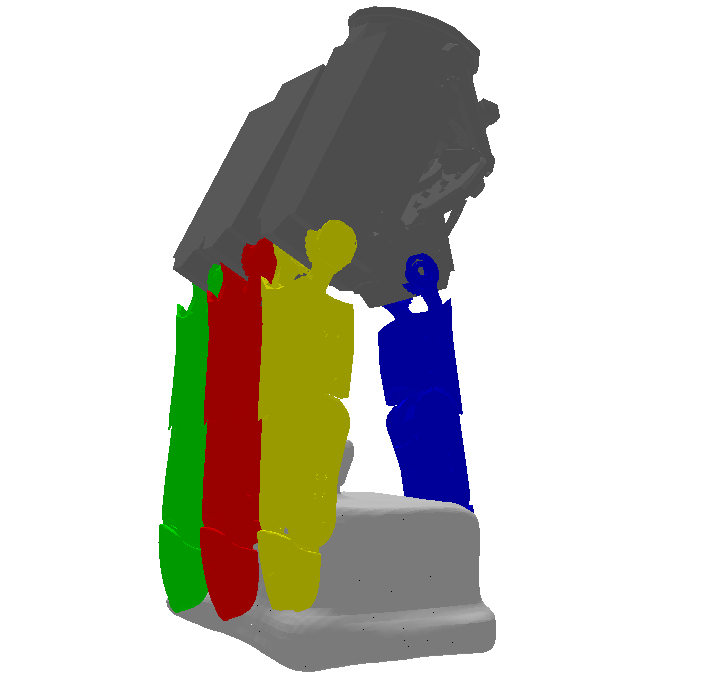}%
}
\subfloat[$s_2 = 5.34$ \\ $c_2 = \SI{6}{\percent}$]{%
  \centering%
  \includegraphics[width=.14\linewidth]{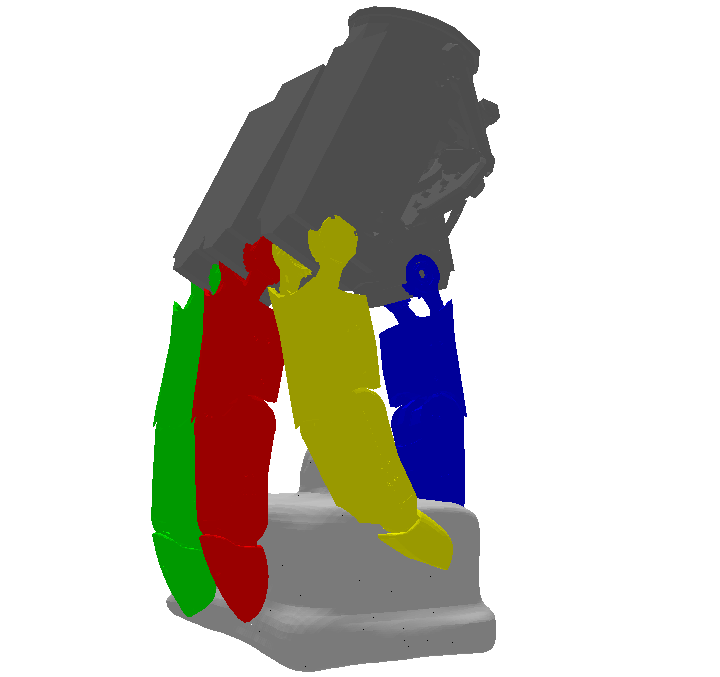} %
}
\subfloat[$s_3 = 5.45$ \\ $c_3 = \SI{39}{\percent}$]{%
  \centering%
  \includegraphics[width=.14\linewidth]{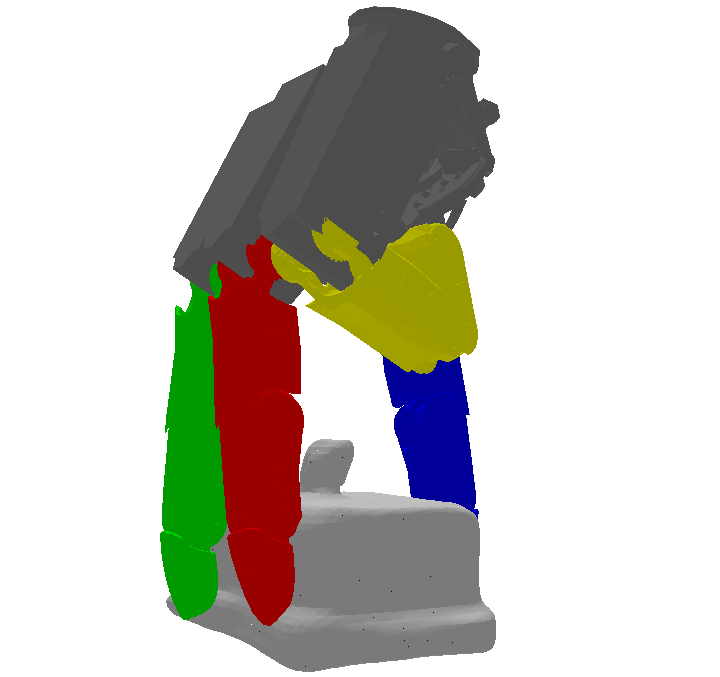}  %
}
\subfloat[$s_4 = 1.26$ \\ $c_4 = \SI{2}{\percent}$]{%
  \centering%
  \includegraphics[width=.14\linewidth]{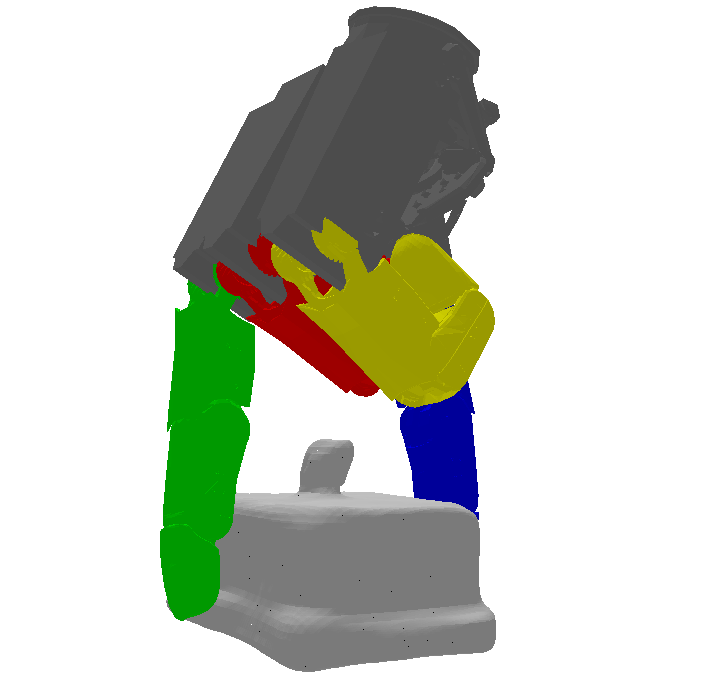}%
}
\caption{Comparing single-output network (left) with multi-output network (right) on predicting the grasp for a butter box: The prediction of the single-output network is a mixture of all modes leading to one finger intersecting the object. The prediction of the multi-output network can explicitly consider several modes. In this case, two modes are used: The second one using all four fingers and the fourth one using only three.}
\label{fig:exp_ambiguity}
\vspace{-10pt}
\end{figure*}

By applying the architectural changes described in \cref{sec:ambiguity} to the grasping network, we are able to train our network using only one label for each training sample.
When training the model architecture used in \citet{Winkelbauer2022GraspPredictor} on the same data, ambiguities cannot be distinguished and the network learns the mean of all valid joint configurations.
This becomes especially clear when there exist distinct modes in the training labels.
\cref{fig:exp_ambiguity} shows an example where the object can be grasped with three fingers or four fingers.
While the multi-output network can predict separate valid modes, the single-output network predicts a mix between the modes leading to one finger intersecting the object.
This example also shows the importance of classifying which predicted modes are active.
For this kind of object, there seem to exist two main types of grasps in the training dataset.
One using four fingers and one using only three.
The network predicts both modes, each having a  large frequency classification, while the other three outputs are not used here and therefore are classified as close to \SI{0}{\percent}.

\section{Conclusion}
In this paper, we have proposed a novel, complete, and fast pipeline (only \SI{1}{\second} from looking to grasping) for robotic grasping of unknown objects with a multi-fingered hand based only on partial shape observations from a commodity depth sensor. Our method combines deep implicit shape completion (\SI{0.7}{\second}) with data-driven multi-finger grasp prediction (\SI{0.3}{\second} for 1000 grasps). This enables efficient grasping across objects with significant shape variation, a key capability for robotic manipulation in unconstrained real-world environments.
Through systematic experiments on synthetic and real data, we have validated the benefits of our training procedures for improved sim-to-real transfer. Robotic grasping trials further demonstrate the utility of our approach by successfully manipulating a diverse range of novel household objects based on single-view depth images.
Future work will include further improvements to sensor simulation, network architectures and training procedures.


\scriptsize
\bibliographystyle{IEEEtranN-modified}
\bibliography{IEEEabrv, bibliography}


\end{document}